\documentclass[11pt,a4paper]{article}
\usepackage[hyperref]{acl2021}
\usepackage{times}
\usepackage{latexsym}
\usepackage{graphicx}

\usepackage{lipsum}
\newcommand\blfootnote[1]{%
  \begingroup
  \renewcommand\thefootnote{}\footnote{#1}%
  \addtocounter{footnote}{-1}%
  \endgroup
}

\usepackage{microtype}

\aclfinalcopy % Uncomment this line for the final submission
\def\aclpaperid{2900} %  Enter the acl Paper ID here

\title{An Exploratory Analysis of the Relation Between \\ Offensive Language and Mental Health}
\author{Ana-Maria Bucur\textsuperscript{1}, Marcos Zampieri\textsuperscript{2}, and Liviu P. Dinu\textsuperscript{1} \\
  \textsuperscript{1}University of Bucharest, Romania\\
  \textsuperscript{2}Rochester Institute of Technology, USA \\
  \texttt{ana-maria.bucur@drd.unibuc.ro, marcos.zampieri@rit.edu} \\
  \texttt{ldinu@fmi.unibuc.ro}}
\date{}

\begin{document}
\maketitle
\begin{abstract}

%With vast amounts of data available on social media, increased efforts have been invested in the use of NLP methods applied to  mental health.

In this paper, we analyze the interplay between the use of offensive language and mental health. We acquired publicly available datasets created for offensive language identification and depression detection and we train computational models to compare the use of offensive language in social media posts written by groups of individuals with and without self-reported depression diagnosis. We also look at samples written by groups of individuals whose posts show signs of depression according to recent related studies. Our analysis indicates that offensive language is more frequently used in the samples written by individuals with self-reported depression as well as individuals showing signs of depression. The results discussed here open new avenues in research in politeness/offensiveness and mental health. 

%efforts have been invested in the use of NLP methods applied to  mental health. 

\end{abstract}

\section{Introduction}
\label{Introduction}

The use of offensive language is pervasive in social media and it has been studied from different perspectives. A popular line of research is the study of computational models to identify offensive content online relying on traditional machine learning classifiers (e.g. naive bayes and SVMs) \cite{xu2012learning, dadvar2013improving}, neural networks (e.g. LSTMs, GRUs) with word embeddings \cite{aroyehun2018aggression, majumder2018filtering}, and more recently, transformer models like ELMO \cite{peters-etal-2018-deep} and BERT \cite{devlin2019bert} which have shown to obtain competitive scores topping the leaderboards in recent shared tasks on offensive language and hate speech detection \cite{liu-etal-2019-nuli}. \blfootnote{WARNING: This paper contains offensive words.}

Offensive language is related to the notion of impoliteness \cite{culpeper2011impoliteness} and it can take various forms from general and often harmless profanity to abusive language intended to cause harm, such as cyberbullying and hate speech \cite{waseem2017understanding}. Computational models have been applied not only to identify the various types of offensive content \cite{basile2019semeval} but also to, for example, study the relation between profanity and hate speech \cite{malmasi2018challenges} and the different functions and intentions of vulgarity in social media \cite{holgate-etal-2018-swear}. 

Most of the datasets used in the aforementioned studies contain data sampled from the general population and therefore very little light has been shed on the use of offensive language in online communication by specific groups such as individuals with mental health conditions. A notable exception is the recent study by \citet{birnbaum2020identifying} which shows that users with mood disorders (bipolar disorder, major depressive disorder) and schizophrenia spectrum disorders use more swear words in their Facebook messages than healthy users. 
%Furthermore, to the best of our knowledge, the use of computational modelling on the interface between offensive language and mental health has not been explored thus far.  

%Marcos, make it clear that showing vs. not-showing refers only to DEPRESSED individuals. 

To address this shortcoming, in this paper, we build on recent work on offensive language identification and apply it to mental health datasets. More specifically, we look at the role of offensive language in the communication of users with depression using two publicly available datasets containing posts by individuals with self-reported depression diagnosis. 

To the best of our knowledge, this study is the first to apply state-of-the-art offensive language identification models to mental health datasets. We aim to answer two research questions:
%\vspace{-3mm}
    \paragraph{RQ1:} Are posts from individuals suffering from depression more likely to contain offensive language in existing datasets? %Self-reported
%\vspace{-3mm} 
    \paragraph{RQ2:} Are there differences in the nature of offensive language used by individuals with depression compared to control groups? %Previous research has investigated Rissola.... 
    %Level B
    
    % \item {\bf RQ3:} Is the use of offensive content in samples from individuals with self-reported depression related to negative sentiment polarity? Recent studies have analyzed at the role of vulgarity in sentiment polarity across different demographics but the analysis of persons with depression has been mostly neglected.
%\end{itemize}

% The main contributions of this paper are the following:

% \begin{itemize}
%     \item 
%     \item
% \end{itemize}

\section{Related Work}
\label{Related Work}

Offensive language identification is a popular topic in NLP. Researchers have been working to improve the performance of systems trained to identify conversations that are likely to go awry \cite{zhang2018conversations} and to detect the various types of offensive posts in social media \cite{basile2019semeval,kumar-etal-2020-evaluating}. More recently, with the goal of improving explainability, offensive language identification at the token-level has received more attention \cite{mathew2020hatexplain,ranasinghemudes}. A number of computational models have been applied to this task ranging from traditional machine learning classifiers, most notably SVMs \cite{macavaney2019hate}, to various deep learning models \cite{liu-etal-2019-nuli}. While the clear majority of studies on this topic deal with English, some studies have addressed offensive language in other languages like Greek \cite{pitenis2020} and Turkish \cite{coltekin2020} while a few others have applied cross-lingual models to take advantage of existing English datasets when making predictions in languages with fewer resources \cite{ranasinghe-etal-2020-multilingual}.

Several studies have applied machine learning and NLP methods to address research questions related to mental health in social media such as identifying users with a particular mental health condition and predicting the risk of self-harm or suicide ideation  \cite{de2013predicting,preoctiuc2015role,malmasi2016predicting,de2016discovering,chancellor2020methods}. 
The CLPsych workshop co-located with international NLP conferences has hosted multiple competitions on these topics providing participants with important benchmark datasets and attracting a large number of teams \cite{coppersmith2015clpsych,milne2016clpsych,zirikly2019clpsych}. 

There have been multiple studies on the impact of offensive and hateful speech on the individual's psychological mental health and well-being \cite{bannink2014cyber,saha2019prevalence}. The use of offensive language by individuals with mental health conditions, however, has not been substantially studies with the exception of \citet{birnbaum2020identifying} that analyzed the use of offensive language in Facebook messages from individuals with mood disorders. Our work fills this important gap by providing further empirical evidence of the use of offensive language by individuals with diagnosed depression or showing signs of depression. 

\section{Data}
\label{Data}

In our experiments, we use three publicly available English datasets with data collected from social media: one with offensive language annotation, and two datasets with posts from users with self-reported depression diagnosis.

\paragraph{Offensive Language} We use the Offensive Language Identification Dataset (OLID) \citep{zampieri2019predicting} to train offensive language identification models. OLID contains a total of 14,100 manually annotated posts from Twitter and it was released as the official dataset of SemEval-2019 Task 6 (OffensEval) \cite{offenseval}. We chose OLID due to its general hierarchical annotation taxonomy with the following levels: \\

\vspace{-4mm}

\noindent {\bf Level A:} Offensive language identification: offensive (OFF) vs. non-offensive (NOT) \\

\vspace{-4mm}

\noindent {\bf Level B:} Categorization of offensive language: targeted insult or threats (TIN) vs. untargeted profanity (UNT).\\

\vspace{-4mm}

\noindent {\bf Level C:} Offensive language target identification: individual (IND) vs. group (GRP) vs. other (OTH).\\

\vspace{-4mm}

\noindent This hierarchical taxonomy provides us with a flexibility as it represents multiple types of offensive content in a single annotation scheme (e.g. posts targeted at an individual are often {\em cyberbullying} and posts targeted at a group are often {\em hate speech}) making it a great fit for this kind of analysis. In our experiments, we consider level A (offensive vs. non-offensive) and level B (target vs. untargeted).

\paragraph{Mental Health} We run all our experiments on the Reddit Self-reported Depression Diagnosis (RSDD) dataset \cite{yates-etal-2017-depression} and on the Early Risk Prediction on the Internet (eRisk) 2018 dataset \cite{losada-crestani2016}, two publicly available datasets containing posts from Reddit. The RSDD dataset consists of users annotated as having depression by their mention of diagnosis and control users, which are users who do not suffer from depression (there is not any mention of diagnosis in their posts). To prevent users labeled with depression to be easily identified by specific keywords, the authors removed posts containing depression terms (e.g. \textit{depression, depressive}) or belonging to mental health related subreddits. The authors made the training, validation, and test splits available and in our experiments we use the training split, which contains over 5 million posts from users with depression and over 30 million posts from users in the control group.

% contains 3,070 users with depression and 35,753 control users,

% The eRisk 2018 contains 125 users annotated as having depression by their mention of diagnosis and 752 control users as training data. The test split contains 79 users with depression and 741 control users. 

The eRisk 2018 dataset contains users labeled with depression by their mention of diagnosis and control users. In this paper, we use both train and test splits, consisting of a total of approximately 90,000 submissions from users annotated as having depression and 985,000 posts and comments from the users in the control group. As opposed to the RSDD dataset, the authors removed only the posts containing the exact mention of diagnosis.

\section{Methods}
\label{Method}

\paragraph{Offensive Language Detection and Categorization} We address \textbf{RQ1} and \textbf{RQ2} by studying the language of users from the two groups, self-reported depression diagnosis and control, in social media. We start by computing an offensive score, which measures the extent to which a post is offensive, and whether it is a targeted insult or an untargeted post (most often profanity). These two tasks correspond to OLID levels A and B respectively \citet{zampieri2019predicting}. 

For the task of offensive language detection, we fine-tune a BERT model on the OLID dataset on level A. We train the model for 2 epochs, with a small learning rate of $0.00001$ and Adam optimizer \cite{kingma2014adam}. We use an 80:20 split of the training data to choose the best performing model in terms of F1 score.  The model obtains 0.85 Precision, 0.74 Recall and 0.77 F1 score on the test data from the OLID dataset. These numbers are consistent with the baselines reported in \cite{zampieri2019predicting}. The offensive score is computed as a probability taken from the softmax output of the BERT model. 

For the task of offensive language categorization (targeted insult or untargeted profanity) we also choose a transformer-based approach, using another BERT model trained on OLID level B. We fine-tune BERT for 7 epochs with the same aforementioned train-validation split, with a learning rate of $0.00002$ with Adam optimizer and a linear warm-up schedule with a 0.05 warm-up ratio, as proposed by \citet{rosenthal2020large}. To account for the class imbalance, we use cross-entropy loss with balanced class weights. The effectiveness of the model is also evaluated on the OLID test data, using the same metrics and achieving 0.78 Precision, 0.84 Recall and 0.80 F1 score. 

% TODO maybe reformulate - not detect, maybe split into two groups
\paragraph{Signs of Depression Detection} Furthermore, we are interested in distinguishing the posts that show signs of depression from all the posts of individuals from the depression group. This way, we filter out the noise added by the texts which do not contain any cues of depression. We are using the Semantic Polarity Score heuristic (H$_{s}$ heuristic) proposed by \citet{rissola2020dataset} to detect posts showing signs of depression written by individuals with a self-reported depression diagnosis. 

H$_{s}$ uses a mix of sentiment polarity, depression score, and emotion detection. 
%to detect the posts with signs of depression. 
The authors use TextBlob\footnote{https://textblob.readthedocs.io/en/dev/index.html} to obtain the polarity score of each post, ranging between -1 and 1. The terms from EmoLex \cite{mohammad2013crowdsourcing} are used in order to detect the emotions (anger, fear, anticipation, trust, surprise, sadness, joy, and disgust) contained in the texts. The depression score of each post is computed using the NRC Affect Intensity Lexicon \cite{mohammad2017word}, ranging from 0 to 1. In order to distinguish the posts showing signs of depression from other posts of users with self-reported depression diagnosis, we follow the criteria from \citet{rissola2020dataset}. Posts are labeled as showing signs of depression if the texts have a negative polarity, if sadness or disgust emotions are present, and if they have a depression score higher than 0.1.

\section{Results and Discussion}
\label{Results}

\begin{figure}[hbt]
    \centering
    \includegraphics[width=1.02\linewidth]{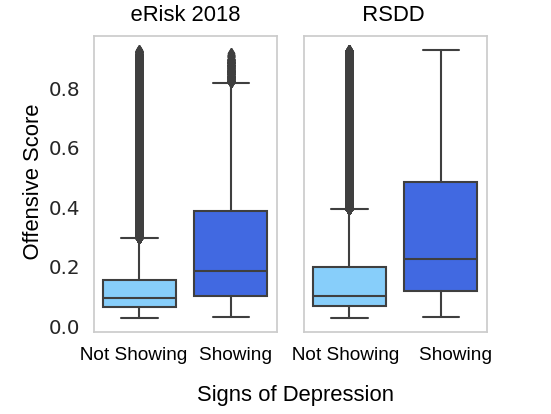}
    \vspace{-8mm}
    \caption{Distribution of the offensive language score for posts written by users with self-reported depression diagnosis and showing or not showing signs of depression measured with the H$_{s}$ heuristic.}
    \label{fig:boxplot_Hs}
\end{figure}

Using the H$_{s}$ heuristic, we demonstrate that there is a statistically significant difference (Welch t-test, $p$-value $<$0.001) in terms of offensive language use between individuals with self-reported depression diagnosis that manifest signs of depression in their posts and users who do not show any signs of depression. Posts containing signs of depression have a higher offensive score than posts from users diagnosed with depression without any signs, in both eRisk 2018 and RSDD datasets, as shown in Figure \ref{fig:boxplot_Hs}. 

% \begin{figure}[hbt]
%     \centering
%     \includegraphics[width=1\linewidth]{acl-ijcnlp2021-templates/images/Distrib.png}
%     \caption{Distributions of the offensive language score for posts of users labeled with depression and control. The differences between the two distributions of offensive score are statistically significant (Welch t-test, $p$-value $<$0.001).}
%     \label{fig:distributions}
% \end{figure}
% maybe remove this figure

% Given the distributions of the offensive score for depression and control (Figure \ref{fig:distributions})

For labeling the offensive posts, we use the same 0.50 threshold as used during training. We show in Table \ref{table:offensive} that more posts from users diagnosed with depression are labeled as offensive than from control. Using the H$_{s}$ heuristic, we filter the posts containing signs of depression and find that there is a higher percentage of posts with signs of depression labeled as offensive. These findings are consistent for both eRisk 2018 and RSDD datasets.

\begin{table}[!ht]
\centering
\resizebox{\linewidth}{!}{

\begin{tabular}{lcc|cc}
\hline & \multicolumn{2}{c}{\textbf{Self-reported}} & \multicolumn{2}{c}{\textbf{Signs of depression}}\\ \hline
\hline \textbf{Dataset} & \textbf{Depression} & \textbf{Control} & \textbf{Showing} & \textbf{Not showing}\\ \hline
\textbf{eRisk 2018} & 8.24\%  & 5.91\%  & 18.50\%  & 7.40\% \\
\textbf{RSDD} & 11.31\%  & 8.91\%  & 24.33\%  & 10.10\% \\
\hline
\end{tabular}
}
\caption{\label{table:offensive} Percentage of posts labeled as offensive from total posts of self-reported individuals and of individuals showing/not-showing signs of depression measured with the H$_{s}$ heuristic.}
\end{table}

\noindent The higher degree to which depressed individuals use offensive language in comparison to individuals in the control group can be explained via the emotion regulation framework \citep{gross1999emotion}. The use of offensive language could be an emotion regulation strategy through which depressed individuals relieve some of their distress. Similarly, pain and distress studies indicate that the use of offensive language when experiencing pain significantly diminishes the level of pain experienced \citep{stephens2020swearing}, suggesting that the use of offensive language can relieve distress.

Although there are more posts with signs of depression labeled as offensive, the majority of them are untargeted (containing swears, profanity) and only 11.48\% and 8.29\%, respectively, are targeted insults (Table \ref{table:target}). 
%Findings from \citet{birnbaum2020identifying} confirm this result: the participants of their study diagnosed with mood disorders (bipolar disorder, major depressive disorder) and schizophrenia spectrum disorders were more likely to use swear words in their Facebook messages than controls.
%Posts showing signs of depression have a lower percentage of targeted insults than posts of users who do not show cues for depression. 

\begin{table}[!ht]
\centering
\resizebox{\linewidth}{!}{
\begin{tabular}{lcc|cc}
\hline & \multicolumn{2}{c}{\textbf{Self-reported}} & \multicolumn{2}{c}{\textbf{Signs of depression}}\\ \hline
\hline \textbf{Dataset} & \textbf{Depression} & \textbf{Control} & \textbf{Showing} & \textbf{Not showing}\\ \hline
\textbf{eRisk 2018} & 24.12\%  & 21.72\%  & 11.48\%  & 26.68\% \\
\textbf{RSDD} & 16.63\%  & 23.94\%  & 8.29\%  & 18.48\% \\
\hline
\end{tabular}
}
\caption{\label{table:target} Percentage of posts labeled as targeted insult from the offensive posts of self-reported individuals and of individuals showing/not showing signs of depression measured with the H$_{s}$ heuristic.}
% \vspace{-10mm}
\end{table}

The fact that depressed individuals tend to use more self-deprecating content and less deprecation of others, as evidenced in our analysis, is a result that is in line with the broad spectrum of cognitive studies, which indicates that negative evaluation of the self is a main interpretation bias in depressed individuals \cite{everaert2017comprehensive}. Depressed individuals tend to view themselves as less valuable than others. By self-deprecating language, we use the definition from \citet{speer2019reconsidering}. This broader definition includes, but is not limited to, insults towards self, if they have a negative intention. Finally, studies show that there is also a self-focused attention tendency in depressed individuals \citep{brockmeyer2015me}, where just like in other conditions (e.g. anxiety), individuals tend to be unable to detach from their own perspective focusing primarily on their side of the story, their pain, etc.

% This finding is consistent with the existing literature: individual diagnosed with depression have a more self-focused language \cite{chung2007psychological, de2013predicting}. Thus, posts with signs of depression are less likely to contain insults targeted on others.

\begin{figure}[!hbt]
    \centering
    \includegraphics[width=1.04\linewidth]{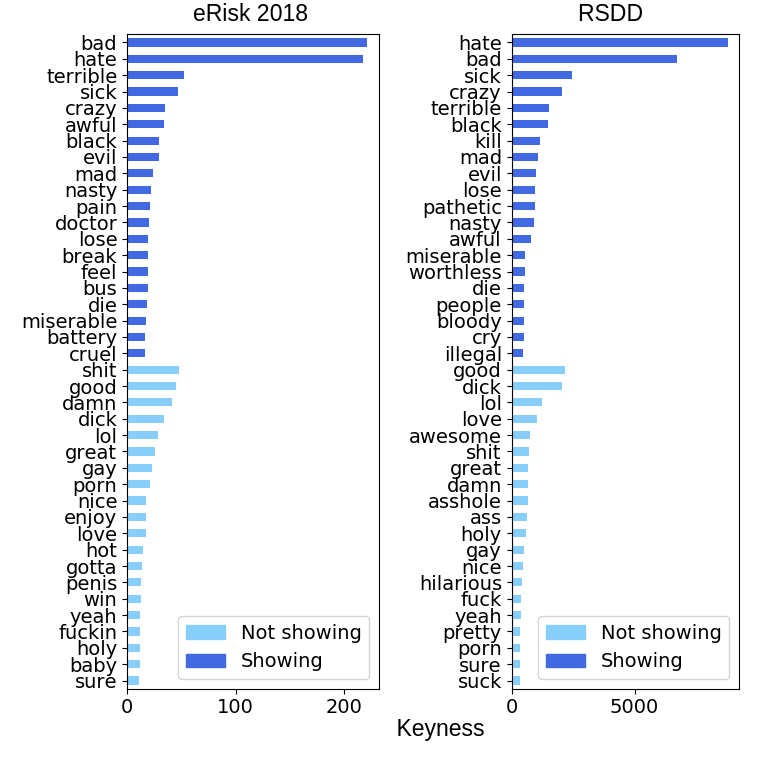}
    \vspace{-10mm}
    \caption{Keyness for words from posts showing/not showing signs of depression.}
    \label{fig:keyness}
\end{figure}

%Marcos reword it. 
In order to further understand the differences in the use of offensive language, we analyze the words from posts written by individuals with depression. We compute the keyness score \cite{kilgarriff2009simple,gabrielatos2018keyness} of content words (removing stop words) from posts labeled as offensive written by users with self-reported diagnosis. 
% We compare the samples written by users showing signs of depression with the samples written by users who do not show any signs. 
The keyness is computed in order to show which words occur more often in the texts from depressed individuals showing signs of depression (target corpus) in comparison to the texts from users diagnosed with depression that do not show signs of depression (reference corpus). We calculate the frequencies of words from the two corpora and then the log-likelihood Ratio (G$^2$) \cite{dunning1993accurate} for each word. In Figure \ref{fig:keyness} we present the top 20 words, ordered by G$^2$ from each corpus, in the two datasets.

We show that, while users without signs of depression refer more to sexual and profane terms, posts by users showing signs of depression include more negative words such as \textit{bad, hate, sick, death}.  This result corroborates the findings described in the literature on cognitive errors or biases in depression \cite{beck2014advances}. It is well known that depressed individuals tend to view life events more negatively than their non-depressed peers \cite{everaert2017comprehensive}. Furthermore, depressed individuals are more likely to recall negative life events than positive events and also more likely to pay closer attention to negative information \cite{beck2014advances}. Signs of this biased view of life are expected to be noticeable in language and there are studies that indicate that depressed individuals tend to have a more negative discourse than their non-depressed depressed peers \cite{rude2004language}. Keywords with a negative polarity, such as \textit{bad, die} or \textit{pain}, seem to be pervasive in the speech of depressed individuals as confirmed in our study. Finally, the reduced sexual drive is a well-known indication of depression \cite{manohar2017sexual}, therefore, it is to be expected that depressed individuals tend to use fewer words with sexual connotation as confirmed in our study. 

\section{Conclusion and Future Work}
\label{Conclusion}

This paper is the first to apply offensive language identification techniques to posts by individuals with a mental health condition with the purpose of interpreting the use of profanity and offensive language by this group. We showed how the offensive language use differs substantially between individuals with depression (in samples with self-reported diagnosis or showing signs of depression) answering our {\bf RQ1}. Our findings indicate that users with self-reported depression diagnosis are more likely to use offensive language in their posts compared to the control group. From the posts of individuals with depression, the ones showing signs of depression contain more offensive language than the ones not showing any signs.

In terms of the nature of offensive content, our results indicate that posts from individuals with signs of depression are less likely to contain targeted offensive language. Furthermore, while analyzing the texts of users with depression, we observed a larger frequency of words with negative polarity (e.g. \textit{bad, hate, sick, suffer}) in the posts of users showing signs of depression, where the discourse of users not showing any signs contains more sexual-related content, addressing our {\bf RQ2}. These findings are consistent with the existing literature from psychology \cite{stephens2020swearing, everaert2017comprehensive, beck2014advances}.

While it is clear that depressed users are more likely to write posts with negative polarity, the interplay between offensive language and polarity in the mental health datasets used in this paper has not yet been explored. A polarity score has been used in the heuristic by \citet{rissola2020dataset} suggesting that using NLP models to investigate the interplay between polarity and depression is a promising future work direction. Other future work directions include the analysis of the targets of offensive posts using the OLID Level C annotation and a more detailed analysis on the function of profanity and vulgarity in these datasets \cite{holgate-etal-2018-swear}. Finally, we would like to carry out a similar analysis for other languages taking advantage of existing datasets and available cross-lingual embedding models. 

\section*{Acknowledgments}

We would like to thank Ioana Podină for providing us with helpful feedback and suggestions for the results analysis. We further thank the anonymous ACL reviewers for the insightful feedback provided. Finally, we thank the dataset creators for making the data used in this study available. 

Liviu P. Dinu was partially supported by a grant of the Ministry of Research, Innovation and Digitization, CNCS/CCCDI – UEFISCDI, project number 108, within PNCDI III.

\section*{Ethics Statement}

This paper uses publicly available datasets on offensive language and mental health to train computational models with the purpose of carrying out both quantitative and qualitative data analysis. We are primarily interested in quantifying and analyzing the use of offensive language in the texts included in the two mental health datasets and we do not attempt to predict mental health status or condition from these datasets. Potential biases in our model predictions and in our analysis may arise from the annotation and sampling techniques of these two datasets and are not intentional. Finally, we did not use any form of demographic information in our models or in our analysis.

\bibliographystyle{acl_natbib}
\bibliography{acl2021}

% \clearpage
% \appendix
% \renewcommand{\thepage}{}
% \section{Appendices}
% \setcounter{figure}{0}
% \subsection{Computing infrastructure and runtime}

% We used two NVIDIA GeForce RTX 2060 GPUs to train the transformer models used in this work. 
% The runtimes for training the BERT model for offensive language detection (Level A from OLID) and for offensive language categorization (Level B from OLID) were around 5 minutes each.

% \subsection{Offensive score threshold}

% % 
% Given the distributions of the offensive score for depression and control assigned by our model (Figure \ref{fig:distributions}), we decide to use a 0.75 threshold for the offensive score in order to filter the posts that are actually offensive.

% \begin{figure}[hbt]
%     \centering
%     \includegraphics[width=1\linewidth]{acl-ijcnlp2021-templates/images/Distrib.png}
%     \caption{Distributions of the offensive language score for posts of users labeled with depression and control. The differences between the two distributions of offensive score are statistically significant (Welch t-test, $p$-value $<$0.001).}
%     \label{fig:distributions}
% \end{figure}

\end{document}